%% file: acs-latex-template.tex
\documentclass[journal=jacsat,manuscript=article]{achemso}

\usepackage[version=3]{mhchem} % Formula subscripts 

\usepackage{chemformula} % Formula subscripts using \ch{}
\usepackage[T1]{fontenc} % Use modern font encodings
\usepackage{adjustbox}
\usepackage{multirow}
\usepackage{booktabs} 
\usepackage{amssymb} 
\usepackage[table]{xcolor}
\usepackage{rotating}
\usepackage{float}
\usepackage{enumitem}
\usepackage[normalem]{ulem}
\usepackage{xcolor}
\usepackage{graphicx}%
\usepackage{amsmath,amssymb,amsfonts}%
\usepackage[table]{xcolor}%
\usepackage{hyperref}
\usepackage{booktabs}%
\usepackage{algorithm}%
\usepackage{algpseudocode}%
\usepackage{multirow}
\usepackage{enumitem}

\newcommand{\mname}{SpecMol}
\newcommand{\mnametwo}{SpecMol-Bench} 
\author{Shuaike Shen}
\affiliation{Carnegie Mellon University, Pittsburgh, PA, United States of America}
\altaffiliation{These authors contributed equally to this work.}
\author{Jiaqing Xie}
\affiliation[PJLAB]{Shanghai Artificial Intelligence Laboratory, Shanghai, China}
\altaffiliation{These authors contributed equally to this work.}
\author{Zhuo Yang}
\affiliation[PJLAB]{Shanghai Artificial Intelligence Laboratory, Shanghai, China}
\alsoaffiliation[XDU]{Xidian University, Xi’an, Shaanxi,China}
\altaffiliation{These authors contributed equally to this work.}
\author{Antong Zhang}
\affiliation{Brown University, Rhode Island, RI, United States of America}
\author{Shuzhou Sun}
\affiliation[PJLAB]{Shanghai Artificial Intelligence Laboratory, Shanghai, China}
\alsoaffiliation
{University of Oulu, Oulu, Finland}
\author{Ben Gao}
\affiliation[PJLAB]{Shanghai Artificial Intelligence Laboratory, Shanghai, China}
\alsoaffiliation[WHU]{Wuhan University, Wuhan, Hubei, China}
\author{Tianfan Fu}
\affiliation[NJU]{Nanjing University, Nanjing, Jiangsu, China}
\alsoaffiliation[PJLAB]{Shanghai Artificial Intelligence Laboratory, Shanghai, China}
\author{Biqing Qi}
\affiliation[PJLAB]{Shanghai Artificial Intelligence Laboratory, Shanghai, China}
\email{qibiqing@pjlab.org.cn}
\author{Yuqiang Li}
\affiliation[PJLAB]{Shanghai Artificial Intelligence Laboratory, Shanghai, China}
\email{liyuqiang@pjlab.org.cn}

\title[SpecMol]{SpecMol: A Unified Framework for Spectroscopy-Grounded Molecular Modeling and Evaluation} 

\abbreviations{IR,NMR,UV}

\mciteErrorOnUnknownfalse
\begin{document}

%%==================================%%
%% Sample for unstructured abstract %%
%%==================================%%
\begin{abstract}
Large language models have emerged as transformative tools in molecular science, demonstrating remarkable potential in molecular property prediction and de novo molecular design. However, their application to spectroscopy remains notably limited, despite its foundational role in experimental molecular characterization and structural validation. Progress in spectroscopy-grounded reasoning has been hindered by the lack of standardized spectral representations and comprehensive evaluation protocols, making cross-study comparisons difficult. 
To bridge this gap, we present a unified framework for spectroscopy-grounded molecular modeling and evaluation. At its core, the SpecMol foundation model integrates spectral interpretation, molecular representation learning, and three-dimensional structure generation within a single interface. Complementing this, we establish SpecMol-Bench as a systematic evaluation protocol encompassing cross-modal tasks: spectra-to-structure elucidation, structure-to-spectra simulation, and SMILES-to-3D conformation generation. Under this unified framework, SpecMol achieves accurate spectra-driven structure elucidation and reproduces experimental nuclear magnetic resonance characteristics with high fidelity. The model also generates chemically valid three-dimensional conformations directly from SMILES strings and consistently outperforms existing general-purpose molecular language models across standardized evaluation metrics.

\end{abstract}

\input{files/introduction}
\input{files/related_works}
\input{files/methods}
\input{files/results}

\input{files/conclusion}
\input{files/contributions}
\input{files/Conflict_Statement}
\input{files/acknowledgement}
\input{files/software}

\bibliography{achemso}
\newpage
% \appendix
% \input{Support_Information}
\input{files/toc}

\end{document}

%% file: files/introduction.tex
%!TEX root = main.tex
\section{Introduction}\label{sec1}

% 提出了第一个问题，就是别人缺少的用LLM解谱。不仅仅是因为别人没做，而是我们关注到了从分子表示到实验验证的鸿沟。简单来说就是，
Large language models (LLMs) have recently emerged as powerful tools in molecular sciences, leveraging cross-task knowledge transfer to advance applications such as property prediction~\cite{ zhang2024chemllm,zheng2025large,wang2025chem}, reaction forecasting~\cite{tharwani2025large,shi2023relm, chen2025reactgpt, zhao2025chemdfm, ko2025reactionreasoner, bran2025chemical, wang2025chem}, and de novo molecule design~\cite{jiang2025chem3dllm, pardo2026molorggpt, bhattacharya2024large, hu2023novo}. Yet a critical disconnect persists between these in silico advances and laboratory practice: definitive molecular identity and structure assignment still rely on experimental spectroscopy and structural confirmation. Nuclear magnetic resonance (NMR) confirms atomic connectivity and stereochemistry, infrared (IR) spectroscopy identifies functional groups, and mass spectrometry (MS) verifies molecular weight and fragmentation patterns, which collectively form the empirical backbone of structural elucidation~\cite{keeler2011understanding,bellamy2013infra,mclafferty1993interpretation}. Crucially, these spectral fingerprints are intrinsically linked to the molecule's three-dimensional (3D) conformation, which ultimately governs its biological interactions and function. Despite this, most current molecular foundation models operate almost exclusively in symbolic spaces (e.g., SMILES~\cite{Weininger1988-oq}), limiting their ability to directly interpret spectral evidence, reason over 3D geometry, or close the Design-Make-Test-Learn (DMTL) cycle. As a result, LLMs rarely participate in routine laboratory workflows that determine whether a proposed structure is consistent with measured spectra. Their utility thus remains restricted to computational abstraction rather than practical laboratory verification. This gap motivates molecular foundation models that can reason directly over experimental observables and geometrically grounded structures.

A fundamental technical barrier to closing this validation loop is not the lack of data, but the absence of a \emph{model-native interface} for continuous scientific measurements. Experimental spectra including $^1$H/$^{13}$C NMR chemical shifts, IR wavenumbers, MS m/z values and 3D atomic coordinates are inherently real-valued, while general-purpose LLMs are trained to model discrete text sequences. Previous research often addresses this mismatch using modality-specific components such as 1D convolutional encoders for spectra~\cite{xiongatomic, hu2024accurate, jin2025nmr}, graph neural networks for 3D conformers~\cite{fang2022geometry, stark20223d}, or specialized peak-picking modules for NMR~\cite{klukowski2018nmrnet}. Although effective in isolated settings, such hybrid pipelines frequently result in a \textit{semantic disconnect} where the LLM reasons over compressed embeddings or post-processed peaks rather than the raw quantitative signals interpreted by chemists, such as a $^{13}$C shift of 161.49 ppm or an interatomic distance of 1.43~\AA. Furthermore, when numerical values are naively serialized as text, standard tokenizers provide tiny inductive bias for treating high-precision numbers as \emph{continuous and comparable quantities} because token boundaries and digit-level fragmentation can be unstable. This undermines fine-grained reasoning that depends on quantitative proximity, such as distinguishing carbonyl ($\sim$170 ppm) from aromatic ($\sim$130 ppm) carbon signals, or detecting subtle geometric violations in 3D conformations. Therefore, a key prerequisite for spectroscopy-centered molecular reasoning is a standardized \emph{textual serialization schema} that (i) makes modality and field types explicit, and (ii) preserves numerical semantics in a language-compatible form, enabling an LLM to directly attend to experimental measurements without relying on modality-specific encoders.

A significant yet frequently overlooked challenge arises from the fragmentation of evaluation protocols across multi-modal molecular tasks. The field currently lacks a unified standard to assess whether a model consistently understands a molecule across symbolic, geometric, and empirical dimensions. Spectral simulation tasks are often evaluated with disparate tolerances and matching strategies for NMR~\cite{klukowski2018nmrnet, xu2025toward, jin2025nmr}, whereas continuous spectra such as IR and MS frequently utilize inconsistent binning resolutions that result in incomparable similarity scores. Similarly, 3D conformation generation is often assessed only by basic validity (e.g., RDKit parsability) while neglecting critical geometric constraints such as steric clashes, bond length violations, or torsional strain~\cite{jiang2025chem3dllm}. This metric heterogeneity creates an inconsistent comparison landscape, which precludes fair benchmarking and hinders progress toward general-purpose molecular foundation models for reliable spectroscopy-driven workflows. To address this deficiency, the development of a unified, chemistry-informed metric suite that enables reproducible, cross-task benchmarking under consistent protocols is essential.

To enable reproducible evaluation in this setting, we establish \textsc{\mnametwo}, a unified benchmark with consistent and chemistry-aware evaluation metrics spanning spectra-to-structure elucidation, structure-to-spectra simulation, SMILES-to-3D generation, molecular question answering, and chemical name conversions. We further develop \textit{\mname}, a reference foundation model that implements the proposed serialization and is trained under the unified paradigm of \textsc{\mnametwo}. \textit{\mname} 
serves not only as a performance-oriented contribution, but also as a controlled baseline to clarify the limitations of existing LLMs when evaluated under consistent, laboratory-grounded protocols. \textit{\mname} introduces a standardized \emph{textual serialization} for spectroscopic signals and atomic coordinates in which modality types are explicitly marked with lightweight XML-like tags such as \texttt{<13C\_NMR>}, \texttt{<SMILES>}, and \texttt{<IUPAC>}. These modalities can also be represented in a standard SDF format. Furthermore, quantitative values are normalized using a fixed-point format to preserve numerical semantics in a language-compatible sequence, enabling direct cross-modal reasoning without modality-specific encoders. Under this unified framework, \textit{\mname} advances spectroscopy-driven molecular reasoning beyond existing molecular language modeling approaches, enabling end-to-end workflows from spectra to validated structures and from molecular strings to physically plausible 3D conformations.

% \begin{figure}[!htb]
%     \centering
%     \vspace{-2mm}
%     \includegraphics[width=0.9\linewidth]{figs/Pipeline_SpecMol.png}
%     \vspace{-2mm}
%     \caption{(a) Multimodal data sources used for training, including chemical literature, molecular strings (SMILES/IUPAC), experimental spectra (NMR, IR, and MS), and 3D conformations. All continuous signals (spectra, coordinates) are converted into standardized textual tokens (e.g., \texttt{<13C\_NMR>78.1241</13C\_NMR>}) via physics-grounded serialization. 
% (b) Three-stage training pipeline: (1) continual pretraining on chemical text and multimodal token sequences; (2) multi-task mixed supervised fine-tuning to align symbolic, geometric, and empirical modalities through bidirectional tasks (e.g., spectra $\leftrightarrow$ SMILES, SMILES $\rightarrow$ 3D, and SMILES $\leftrightarrow$ IUPAC Names); and (3) instruction tuning with structured templates to enable robust, format-consistent generation for unified evaluation.  Representative applications: MS spectra-to-SMILES structure elucidation, SMILES-to-3D conformation generation, and IUPAC-to-SMILES translation. See Section \textit{Three-phase Learning Strategy} and \textit{Spectrum Textual Description} for details.}

%     \label{fig:pipeline}
%     \vspace{-2mm}
% \end{figure}

\begin{figure}[!htb]
    \centering
    \includegraphics[width=0.9\linewidth]{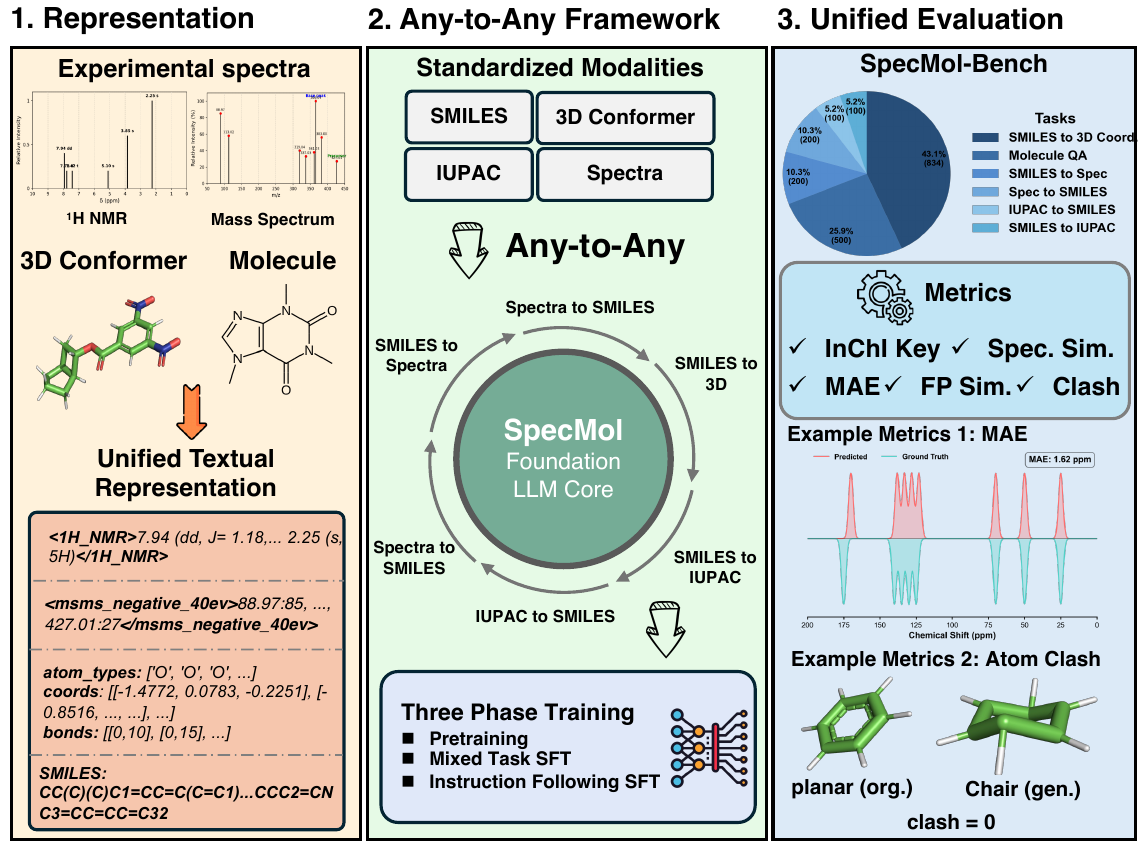}
    \caption{We serialize experimental spectra, 3D conformer's coordinates, and molecule strings into a standardized, numerically faithful textual schema with explicit modality tags, enabling a single LLM to perform any-to-any cross-modal generation and to be evaluated under the unified SpecMol-Bench metric suite.}
    \label{fig:pipeline}
\end{figure}

%% file: files/related_works.tex
\section{Related Works}

\subsection{Spectroscopy-Grounded Models}
Learning-based methods have increasingly connected experimental spectra with molecular structure across NMR, MS, and IR/Raman~\cite{xue2025nmrmind, zhou2025nmrformer, klukowski2018nmrnet, lu2025vib2mol}. 
In the field of NMR, approaches range from retrieval/optimization pipelines such as NMRSolver~\cite{jin2025nmr} to neural or language-model-driven structure elucidation and simulation~\cite{xue2025nmrmind, hu2024accurate, zhou2025nmrformer}. 
Regarding MS, diffusion-based spectrum modeling (e.g., DiffMS~\cite{pmlr-v267-bohde25a}) and specialist tandem-MS elucidation systems (e.g., ICEBerg~\cite{wang2025neural}) have shown strong performance, while waveform-like spectra are also modeled by modern generative frameworks (e.g., DiffRaman~\cite{yao2025difframan}, DiffSpectra~\cite{wang2025diffspectra}). 
Prompt-based large language model formulations such as SpectraLLM~\cite{su2025language} further explore the utility of general-purpose models for spectroscopy-related tasks. Despite rapid progress, results remain difficult to compare across modalities due to inconsistent splits, preprocessing, and evaluation metrics (e.g., tolerances and binning resolutions). We address these inconsistencies with \textsc{\mnametwo}, which provides unified splits, protocols, and cross-modal task coverage.

\subsection{Molecular Foundation Models}
Molecular foundation models have emerged as powerful tools for property prediction and \textit{de novo} design. Early efforts including ChemLLM~\cite{zhang2024chemllm}, ChemMLLM~\cite{tan2025chemmllm}, and NatureLM~\cite{xia2025naturelm} adapt large language models to string-based chemical representations such as SMILES and IUPAC. These models have demonstrated strong performance on symbolic reasoning benchmarks.
Other lines of work incorporate 3D signals to improve molecular representations, e.g., GraphMVP~\cite{DBLP:conf/iclr/LiuWLLGT22} and GeoMol~\cite{DBLP:conf/nips/GaneaPCBJGJ21}. However, most foundation models still treat experimental spectra as external to the core modeling interface, leaving a gap between text-only generation and laboratory validation. 
In this work, \textit{\mname} treats spectra as first-class inputs within a unified language-model interface, including IR/NMR/MS with molecular strings, and 3D structure generation.

\subsection{Cross-Modal Alignment and Evaluation Protocols}
Progress in spectroscopy-grounded molecular learning is also blocked by inconsistent alignment and evaluation protocols. 
For NMR, studies vary widely in peak matching tolerances and matching strategies~\cite{jin2025nmr, xu2025toward}. 
For other spectral modalities, evaluation often depends on binning and preprocessing choices that differ across benchmarks (resolution, smoothing, filtering, normalization), leading to incomparable similarity scores~\cite{zou2023deep, xu2025pretrained}. 
For 3D generation, common metrics emphasize parsability or geometric similarity while often underweighting chemical and physical plausibility (e.g., clashes and bond violations)~\cite{stark20223d, fang2022geometry}; moreover, RMSD-style conformer metrics (e.g., COV/MAT in GeoDiff~\cite{xu2022geodiff} and TorsionDiff~\cite{jing2022torsional}) can be confounded when models produce non-parsable outputs, making comparisons sensitive to validity filtering. To facilitate fair cross-modal evaluation, we provide a chemistry-aware metric suite with standardized spectral preprocessing and geometry-aware validity checks to ensure consistent assessment across spectra, strings, and 3D structures.

%% file: files/methods.tex
\section{Method} \label{sec:methods}

We introduce \textit{\mname} molecular foundation model which integrates spectroscopy-grounded reasoning with 3D molecular structure generation (Fig.~\ref{fig:pipeline}). 
Unlike prior approaches that operate primarily on SMILES, \textit{\mname} treats experimental spectra including nuclear magnetic resonance, infrared, and mass spectrometry, as well as molecular strings and three-dimensional coordinates, as interoperable modalities within a single language-model interface. The key design is a \emph{numerically faithful textual serialization} that represents continuous measurements with explicit modality tags and fixed-point numbers. This approach enables direct sequence-to-sequence generation across modalities without the need for modality-specific encoders.

\subsection{Spectrum Textual Serialization}
\label{sec:spectra_token}

Experimental spectra and 3D coordinates are real-valued signals, whereas LLMs operate on discrete sequences. 
Rather than modifying the tokenizer or introducing separate neural encoders, we convert spectral signals into a \emph{structured text schema} that (i) explicitly marks modality and fields, and (ii) preserves numerical semantics via fixed-point formatting. 
This representation is interpretable, easy to parse, and compatible with standard LLM decoding.

\paragraph{Why fixed-point numbers.}
Directly writing floating-point values can lead to unstable sub-token fragmentation under common byte pair encoding tokenizers. For instance, the conversion of 161.4878 into fragmented tokens such as 161, the decimal point, 48, and 78 weakens the ability of a model to learn proximity relations between close measurements. We therefore format all continuous values using fixed-point strings with consistent decimal places to improve numerical stability and support distance-sensitive reasoning such as neighboring chemical shifts in $^{13}$C NMR. We use a fixed number of decimal places per modality (e.g., 2–4), chosen to match instrument-level resolution; see SI.

\paragraph{Modality-specific schemas.}
We design compact textual formats tailored to each spectrum type.
For $^{13}$C NMR, we serialize chemical shifts enclosed by $\langle\text{13C\_NMR}\rangle$ and $\langle\text{/13C\_NMR}\rangle$.  For example, a $^{13}$C NMR spectrum is represented as
\texttt{<13C\_NMR>(freq=125Hz, solvent=CDCl3) 161.49 130.02 77.16 ... </13C\_NMR>}.
For $^{1}$H NMR, we additionally encode multiplicity, integration, frequency, and solvent; \textbf{centroid} denotes chemical shift (ppm), \textbf{shape} specifies multiplicity (e.g., s, d, t, dd), the $J$-field records coupling constants (Hz), and \textbf{nH} denotes proton count derived from integration. The sequence is enclosed by $\langle\text{1H\_NMR}\rangle$ and $\langle\text{/1H\_NMR}\rangle$.
For waveform-like spectra such as IR (and similarly Raman/UV when available), we optionally apply light denoising and interpolation to standardize resolution (details in SI), then encode the cleaned spectrum as value ranges and frequency--intensity pairs enclosed by $\langle\text{IR}\rangle$ and $\langle\text{/IR}\rangle$. 
For MS/MS, we include acquisition mode and collision energy in the tag name, and represent each peak as an $m/z$ value with relative abundance (e.g., $\langle\text{msms\_negative\_40ev}\rangle \cdots \langle\text{/msms\_negative\_40ev}\rangle$).

By standardizing spectra into structured sequences, \textit{\mname} can directly condition on and generate spectral evidence in a consistent format. 
Different modalities provide complementary constraints where infrared captures functional groups~\cite{stuart2004infrared}, nuclear magnetic resonance resolves local connectivity and stereochemistry~\cite{claridge2016nmr}, and mass spectrometry provides molecular weight and fragmentation signatures~\cite{gross2017mass}. These modalities jointly guide chemically consistent predictions.
% Different modalities provide complementary constraints: IR captures functional groups~\cite{stuart2004infrared}, NMR resolves local connectivity and stereochemistry~\cite{claridge2016nmr}, and MS provides molecular weight and fragmentation signatures~\cite{gross2017mass}, jointly guiding chemically consistent predictions.

\subsection{3D Structure Generation}
\label{sec:3d_generation}

Molecules are inherently three-dimensional, whereas SMILES mainly encodes 2D connectivity with limited stereochemical information. 
To support geometry-aware modeling, \textit{\mname} is trained to generate 3D structures in addition to spectra-conditioned reasoning. We serialize 3D conformers as sequences that include atom types, bond connectivity, and atomic coordinates. Similar to spectra, coordinates $(x,y,z)$ are written as fixed-point numbers to preserve precision and parsing stability. 
To remove permutation ambiguity, we align the coordinate order with the atom order derived from the corresponding canonical SMILES, enabling the model to learn geometry together with connectivity. This design allows \textit{\mname} to generate physically plausible conformations and capture stereochemistry that is critical for reactivity and interactions~\cite{Platzer2025, Baillif2023, Huang2022, Zhang2023}.

\paragraph{Unified Multimodal Modeling.}
With spectra, molecular strings, and 3D structures represented as standardized sequences, we formulate multimodal molecular learning as a unified sequence modeling problem. 
The model supports uni/bidirectional generation across modalities (e.g., spectra $\leftrightarrow$ SMILES, SMILES $\rightarrow$ 3D, and SMILES $\leftrightarrow$ IUPAC) under a single interface without modality-specific encoders.

\subsection{Three-phase Learning Strategy}
\label{sec:training}

Training proceeds in three phases from general chemical knowledge to multimodal alignment and robust structured generation.

\paragraph{Phase 1: Continual Pre-training.}
We initialize \textit{\mname} from the Qwen2.5-7B base model~\cite{qwen2025qwen25technicalreport}. We first perform continual pre-training on a large corpus of publicly available chemistry literature ($\approx$ 10M documents) and then construct a multimodal corpus by integrating molecular properties and structure data from PubChem~\cite{kim2023pubchem}, simulated spectra from QM9S~\cite{zou2023deep}, and experimental spectra from the Multimodal Spectroscopic Dataset~\cite{alberts2024unraveling}. All data are converted to the standardized serialization described in the previous section.

\paragraph{Phase 2: Multi-task Mixed Supervised Fine-tuning (SFT).}
We perform mixed SFT over diverse tasks:
\begin{itemize}
    \item \textbf{Molecule QA}: multiple-choice or short-form questions about molecular properties.
    \item \textbf{3D Structure Generation}: generate coordinates and connectivity from SMILES/IUPAC using the fixed-point 3D serialization.
    \item \textbf{Name Conversion}: bidirectional translation between IUPAC and SMILES with canonicalization.
    \item \textbf{Structure Elucidation}: generate SMILES conditioned on spectra under the tagged serialization.
    \item \textbf{Forward Spectrum Generation}: generate standardized spectra from molecular inputs for fidelity evaluation.
\end{itemize}

\paragraph{Phase 3: Instruction-following SFT (LoRA).}
Domain fine-tuning can sometimes degrade the ability of a model to follow strict formatting~\cite{qi2024aligned, lyu2024prompt}. In our setting, this appears as extra conversational text or malformed tags, which hinders programmatic parsing of outputs.
We therefore use parameter-efficient fine-tuning with LoRA~\cite{hu2022lora}, updating only low-rank adapters to preserve general instruction behavior while improving adherence to structured schemas~\cite{biderman2024lora}. 
We additionally provide \emph{structured task templates} that explicitly specify required tags and fields (e.g., $\langle\text{13C\_NMR}\rangle(\text{freq, solvent})\,\delta:\,\text{shifts}\langle\text{/13C\_NMR}\rangle$), encouraging clean and machine-readable generations.

To prevent data leakage, all instruction-tuning examples are strictly disjoint from evaluation sets.

%% file: files/results.tex
\section{Results}
%%%%%%%%%%%%%%%%%%%%%%%%%%%%%%%%%%%%%%%%%%%%%%%%%%%%%%%%%%%%%%%%%%%%%%%%%%%%%%%%%%%%%%%%%%%%%%%%%%%%%%%%%%%%%%%%%%%%%%%%%%%%%%%%%%%%%%%%%%%%%%%%%%%%%%%%%%%%%%%%%%%%%%%%%%%%%%%%
\subsection{SpecMol-Bench Setup}

\noindent\textbf{Tasks and Data Split.}
We evaluate \textit{\mname} on a suite of downstream tasks spanning molecular question answering, symbolic translation between SMILES and IUPAC names, spectroscopy grounding between SMILES and spectra, and three-dimensional structure generation from SMILES or IUPAC strings. The task-wise training and testing distributions for supervised fine-tuning are summarized in Table~\ref{tab:sft_split}. For three-dimensional coordinate generation, we use a combined set comprising 8,329 training and 834 testing samples.
To prevent scaffold-level leakage in all structure-conditioned settings, we adopt a Bemis--Murcko scaffold split~\cite{bemis1996properties}.

% --- Table: SFT Data Split ---
\begin{table}[!htb]
\centering
\caption{Task-specific training and testing data distribution for supervised fine-tuning (SFT).}
\label{tab:sft_split}
\begin{tabular}{lcc}
\toprule
\textbf{Task} & \textbf{Train Samples} & \textbf{Test Samples} \\ \midrule
Molecule QA & 14,994 & 500 \\
SMILES to IUPAC & 13,727 & 100 \\
IUPAC to SMILES & 16,987 & 100 \\
SMILES to 3D Coord. & 8,329 & 834 \\
SMILES to Spec & 15,653 & 200 \\
Spec to SMILES & 12,000 & 200 \\ \bottomrule
\end{tabular}
\end{table}

\noindent\textbf{SpecMol-Bench.}
We curate \textbf{SpecMol-Bench} as a unified evaluation benchmark for spectroscopy-centered molecular tasks under consistent splits and protocols. SpecMol-Bench aggregates the supervised fine-tuning tasks shown in Table~\ref{tab:sft_split} into a standardized multi-task setting. This benchmark comprises 81,690 training instances and 1,834 test instances in total.\footnote{The benchmark aggregates multiple tasks; a single molecule may contribute instances to more than one task. Detailed construction and prompts are provided in the Supporting Information.}

\noindent\textbf{Pretraining Data.}
We continually pretrain \textit{\mname} with a large chemistry corpus and a multimodal molecular collection spanning molecular text, structures, and spectra. 
PubChem provides the primary source of molecular descriptions. 
To incorporate spectroscopy signals, we integrate experimental NMR from NMRBank~\cite{wang2025nmrextractor} and simulated spectra from QM9S, together with additional multimodal spectroscopic records from the Multimodal Spectroscopic Dataset~\cite{alberts2024unraveling}. 
The data composition is summarized in Table~\ref{tab:pretrain_data}. 
We apply standard quality control measures including the removal of salts and solvents, element and atom count constraints, deduplication by canonical SMILES, and the removal of any overlap with evaluation sets.
After filtering, we obtain $\sim$5M molecules and $\sim$0.2M spectrum--molecule pairs for pretraining.

% --- Table: Pretrain Data ---
\begin{table}[!htb]
\centering
\caption{Summary of pretraining data sources, spectrum classes, and types.}
\label{tab:pretrain_data}
\small
\setlength{\tabcolsep}{4pt}
\begin{tabular}{lccccccc}
\toprule
\multirow{2}{*}{\textbf{Source}} & \multicolumn{4}{c}{\textbf{Spectrum Classes}} & \multicolumn{2}{c}{\textbf{Spectrum Type}} & \multirow{2}{*}{\textbf{Distribution}} \\ \cmidrule(lr){2-5} \cmidrule(lr){6-7}
 & \textbf{NMR} & \textbf{IR} & \textbf{MS} & \textbf{UV-Vis} & \textbf{Exp.} & \textbf{Sim.} & \\ \midrule
PubChem & / & / & / & / & / & / & 11,900,000 \\
QM9S & $\times$ & \checkmark & \checkmark & \checkmark & $\times$ & \checkmark & $\sim$130,000 \\
NMRBank & \checkmark & $\times$ & $\times$ & $\times$ & \checkmark & $\times$ & 225,809 \\
MultiModal & \checkmark & \checkmark & \checkmark & $\times$ & $\times$ & \checkmark & 790,000 \\ \bottomrule
\end{tabular}
\end{table}

\noindent\textbf{Baselines.}
We compare \textit{\mname} against three categories of baselines.
(1) \textbf{Generalist LLMs}: DeepSeek-V3~\cite{deepseekai2024deepseekv3technicalreport}, Qwen3-235B~\cite{qwen2025qwen25technicalreport}, Kimi-K2~\cite{kimiteam2025kimik2openagentic}, OpenAI o3~\cite{openai2025o3}, Gemini-2.5-Flash~\cite{geminiteam2025flash}, and GPT-5~\cite{openai2025gpt5}. 
(2) \textbf{Chemistry-specialist models}: \textit{ChemDFM-v2.0-14B}~\cite{zhao2025chemdfm} and \textit{InternS1-235B}~\cite{bai2025intern}. 
(3) \textbf{Task-specific tools}: STOUT~\cite{rajan2021stout} and OPSIN~\cite{lowe2022opsin} for SMILES$\leftrightarrow$IUPAC conversion.
For forward NMR simulation, we additionally report MestReNova~\cite{willcott2009mestre} where applicable.

\noindent\textbf{Implementation Details.}
\textit{\mname} is initialized from the \textbf{Qwen2.5-7B}~\cite{qwen2025qwen25technicalreport} base model. 
For all LLM baselines, we use a unified prompt template per task and fixed decoding parameters for reproducibility (temperature $=0$, top-$p=1$; max generation length and stop sequences are task-specific but identical across models). 
All outputs are post-processed with the same parser. Invalid or non-parsable generations (e.g., malformed tags or invalid SMILES) are counted as incorrect for exact-match metrics. 
For SMILES generation, we apply RDKit canonicalization before exact-match scoring and additionally report InChIKey-based accuracy. Full decoding and post-processing settings are provided in the Supporting Information.

\subsubsection{Spectroscopy-Grounded Evaluation Metrics}
\label{sec:evaluation}

\begin{figure}[!htb]
    \centering
    \includegraphics[width=0.8\linewidth]{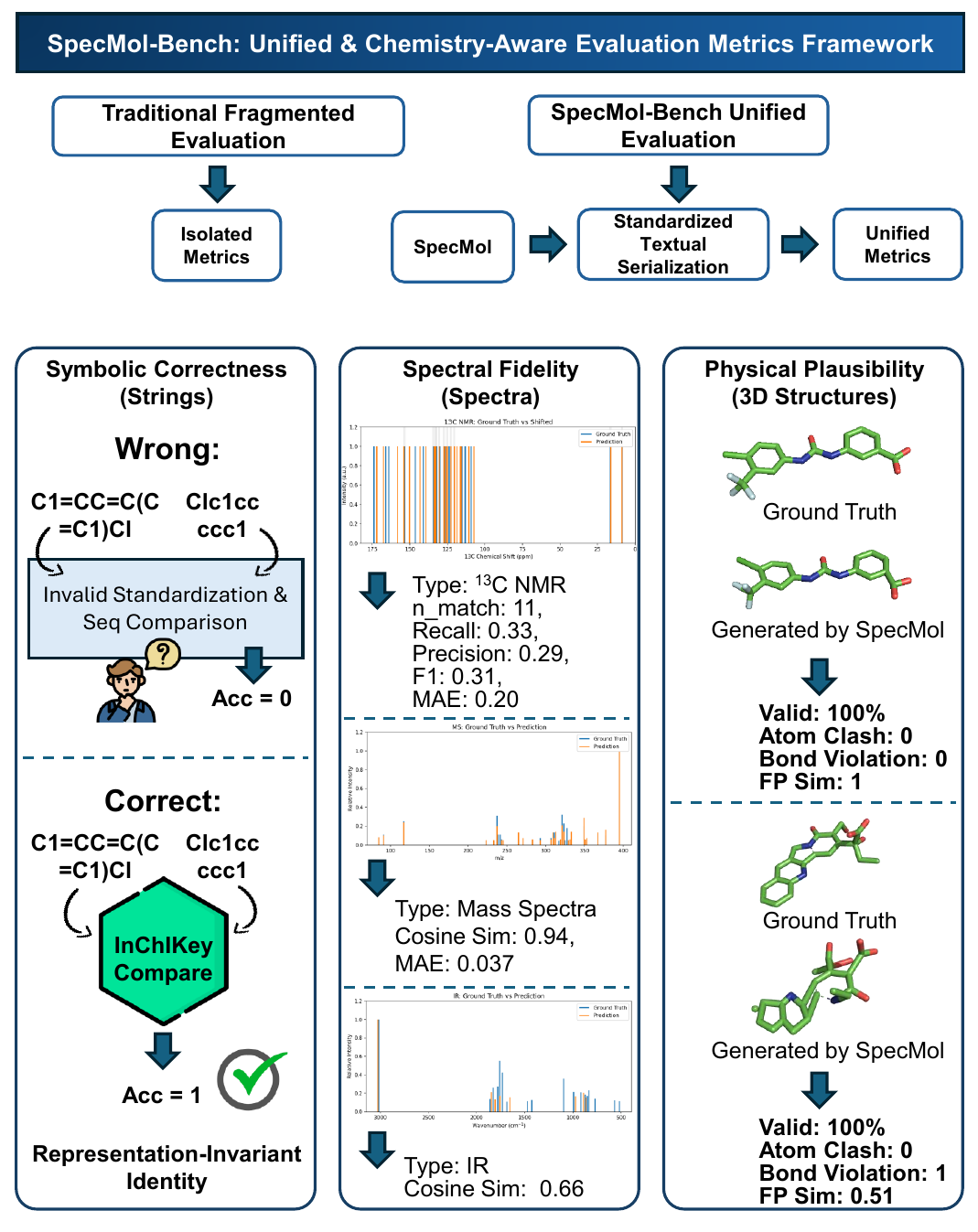}
    \caption{ SpecMol-Bench: A unified evaluation framework. Unlike traditional fragmented protocols, SpecMol-Bench evaluates performance using standardized metrics along three axes: symbolic correctness (strings), spectral fidelity (spectra), and physical plausibility (3D structures).}
    \label{fig:SpecMol_Bench}
\end{figure}

We evaluate performance along three axes: \textbf{(i) symbolic correctness} for molecular strings, \textbf{(ii) spectral fidelity} for spectrum generation, and \textbf{(iii) physical plausibility} for 3D structures (Figure~\ref{fig:SpecMol_Bench}). 
Below we summarize the metrics; complete definitions and implementation details are provided in the Supporting Information. Standardized preprocessing is applied to all models identically

\paragraph{Multiple-Choice Accuracy.}
For molecule QA, we report multiple-choice accuracy to determine whether the predicted option matches the ground-truth label.

\paragraph{Token and Sequence level Accuracy.}
For sequence generation tasks such as SMILES-IUPAC conversion, \textit{token accuracy} measures the fraction of correctly predicted tokens and \textit{sequence accuracy} measures exact-string matches. We additionally report an InChIKey-based accuracy by converting predicted and reference molecules to InChIKeys. A prediction is counted as correct if the InChIKeys match, which provides a representation-invariant measure for SMILES equivalence.

\paragraph{3D Structure Validity and Geometry.}
For 3D coordinate generation, we report:
(i) \textit{SDF Validity}, the percentage of parsable molecules;
(ii) \textit{Atom Clash}, the average number of severe steric overlaps;
(iii) \textit{Bond Violation}, the average number of abnormal bond lengths; and
(iv) \textit{Fingerprint Similarity}, computed as Tanimoto similarity over RDKit fingerprints (path-based, topological torsion, and atom-pair) between predicted and reference structures.

\paragraph{NMR Spectrum Assessment.}
For discrete NMR peak sets, we adopt a tolerance-based peak matching protocol.
For $^{13}$C NMR, we perform greedy one-to-one matching between predicted and ground-truth shifts within a tolerance of 0.5 ppm. We report precision (P), recall (R), F1 score, and mean absolute error over matched pairs. For $^{1}$H NMR, we match peaks within a tolerance of 0.12 ppm and report a weighted Jaccard similarity that jointly reflects shift accuracy and integration fidelity. Weights are defined by a Gaussian decay of the shift difference and scaled by the minimum proton count of the matched pair. We also benchmark unweighted set metrics and the vector cosine similarity score used in NMR-Solver~\cite{jin2025nmr}.

\paragraph{IR and MS Spectrum Assessment.}
For infrared and mass spectrometry spectra treated as continuous intensity distributions, we discretize spectra into vectors via binning at a resolution of 1~cm$^{-1}$ for IR and 1~m/z for MS. We report cosine similarity between predicted and ground-truth vectors to measure global spectral-shape and relative-intensity alignment.

\medskip
\noindent Together, these modality-specific protocols provide standardized and chemistry-aware assessment across spectra, strings, and 3D structures.

%%%%%%%%%%%%%%%%%%%%%%%%%%%%%%%%%%%%%%%%%%%%%%%%%%%%%%%%%%%%%%%%%%%%%%%%%%%%%%%%%%%%%%%%%%%%%%%%%%%%%%%%%%%%%%%%%%%%%%%%%%%%%%%%%%%%%%%%%%%%%%%%%%%%%%%%%%%%%%%%%%%%%%%%%%%%%%%%

\begin{figure}
    \centering
    \includegraphics[width=0.8\linewidth]{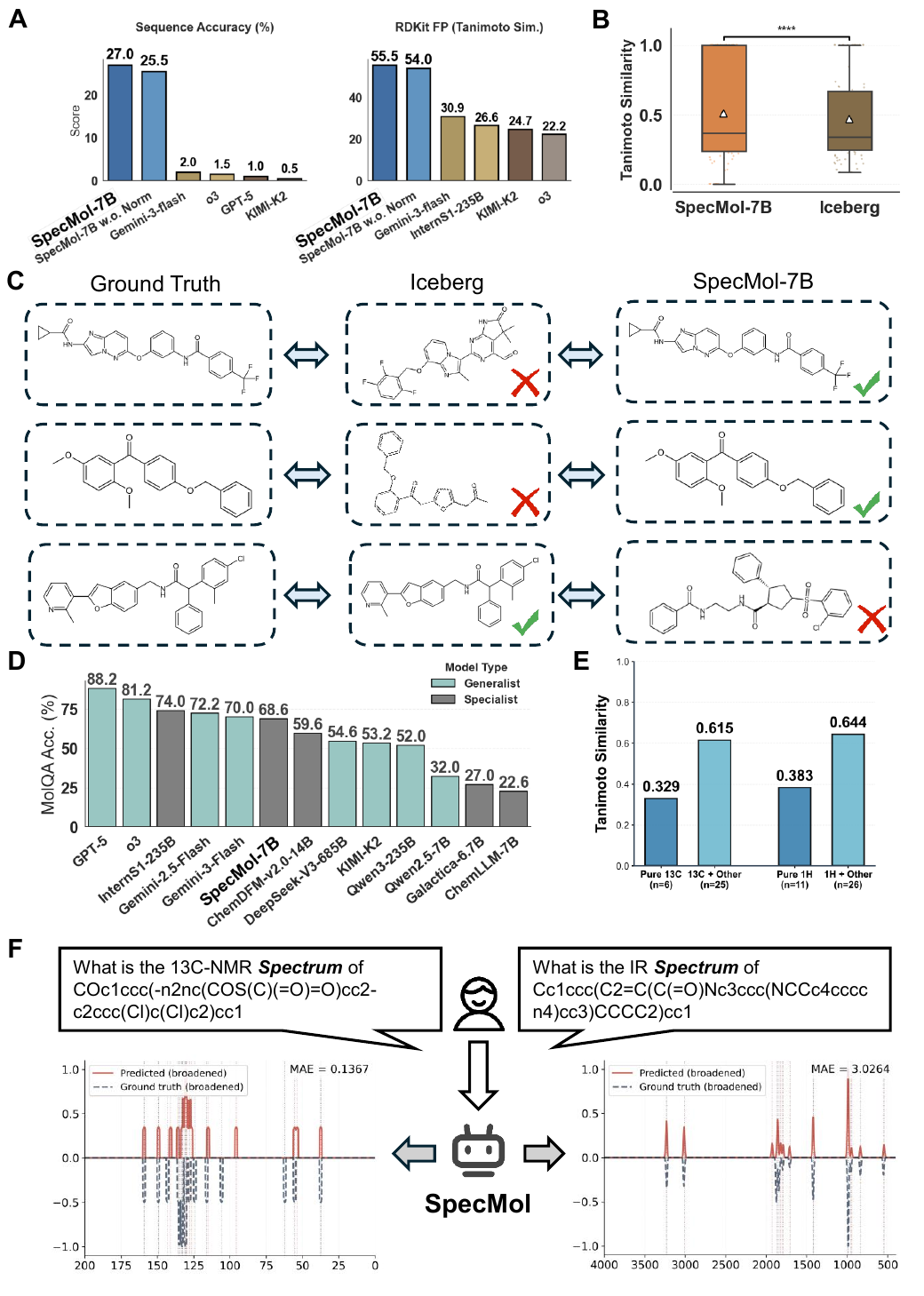}
    \caption{(A-C, E) Results on the spectra-to-SMILES task with evaluation metrics including sequence accuracy, FP similarity, ablation on NMR, and comparison with ICEBerg model~\cite{wang2025neural}  on MS data. (D) Results on MolQA tasks. (F) Case Studies on the SMILES-to-spectra task.}
    \label{fig:performance_overall}
\end{figure}

\subsection{Reverse Problem: Spectra-to-SMILES Molecular Elucidation}
\label{sec:results_elucidation}

We first evaluate \textit{\mname} on the reverse problem of molecular elucidation, which translates multi-modal experimental spectra into SMILES (Table~\ref{tab:spectra_to_smiles}). 
All models are evaluated under a unified protocol with standardized SMILES post-processing (RDKit canonicalization) and fixed decoding settings shared across baselines (see Supporting Information). 
Under this consistent setting, \textit{\mname} achieves the best performance among evaluated specialist and generalist LLM baselines, substantially improving both exact-match accuracy and near-miss structural fidelity.

\paragraph{Large gains over generalist and specialist baselines.}
In Table~\ref{tab:spectra_to_smiles}, the strongest generalist baselines including Gemini-3-flash and o3-mini reach only 2.01\% and 1.50\% sequence accuracy, respectively. In contrast, \textit{\mname} attains a sequence accuracy of 27.00\% which represents a 13-fold to 18-fold improvement. Importantly, \textit{\mname} also improves near-miss structural quality. It achieves an RDKit fingerprint similarity of 55.55\% which nearly doubles the best generalist score of 30.91\% achieved by Gemini-3-flash. This performance indicates that even when the predicted SMILES is not an exact match, the recovered topology remains chemically closer to the ground truth than that of other models. Furthermore, strong specialist baselines such as InternS1-235B do not produce any exact matches under this protocol. This result highlights the difficulty of faithful structure recovery from raw spectral evidence for text-only generalist or single-task systems.
\paragraph{Comparison to a domain expert on MS-only elucidation.}
To validate performance against a dedicated MS elucidation model, we further compare \textit{\mname} with \textbf{ICEBerg}~\cite{wang2025neural} on a curated MS-only test subset (\#104) from SpecMol-Bench. \textit{\mname} achieves a Top-1 accuracy of \textbf{23.08\%}, outperforming ICEBerg (15.38\%). Despite being a unified foundation model rather than an MS-specific expert, \textit{\mname} benefits from broad chemical pretraining and multi-task alignment, enabling stronger generalization even in single-modality settings.

\paragraph{Multi-modal spectra provide complementary constraints.}
A key driver of the performance gains is the synergy between modalities that impose orthogonal structural constraints. As quantified in Figure~\ref{fig:performance_overall}E, using only a single nuclear magnetic resonance modality yields limited recovery with Tanimoto similarities of \textbf{0.329} for $^{13}$C and 0.383 for $^{1}$H. Adding complementary modalities such as infrared and mass spectrometry substantially improves similarity to \textbf{0.615} and \textbf{0.644} respectively. This trend demonstrates that multi-modal evidence effectively reduces ambiguity in structure elucidation.

\paragraph{Serialization improves robustness; remaining failures reflect task-specific inductive biases.}
Fig.~\ref{fig:performance_overall}C provides representative case studies contrasting \textit{\mname} with ICEBerg. \textit{\mname} more reliably reconstructs complex scaffolds and functional groups in drug-like molecules such as trifluoromethyl phenyl motifs, benzophenone ether linkages, and thiophene-pyridinone hybrids. Conversely, ICEBerg often misses the global skeleton or proposes unrelated motifs. Meanwhile, we observe a small set of challenging cases including certain fused heterocycles and benzofuran-like derivatives where ICEBerg succeeds but \textit{\mname} deviates. This suggests that mass-spectrometry-specific fragmentation rules can still provide an advantage for specific chemotypes. Finally, an ablation study shows that applying spectral serialization by adding fixed-point normalization (and standardized tags) yields a consistent gain. Specifically, the SpecMol-7B model without normalization improves from 25.50\% to 27.00\% sequence accuracy in the final model. These results support the importance of standardized numerical representations for stable cross-modal alignment.

\begin{table}[!htb]
\centering
\caption{Performance comparison on the Spectra-to-SMILES translation task (structure elucidation). We compare \textbf{SpecMol-7B} against task-specific specialist models and state-of-the-art generalist LLMs.}
\label{tab:spectra_to_smiles}
\small % 缩放字体大小
\setlength{\tabcolsep}{3.5pt} % 压缩列间距，省出空间防止超宽
\begin{tabular}{lccccc}
\toprule
\textbf{Model} & \textbf{Token Acc.} & \textbf{Seq Acc.} & \textbf{RDK FP} & \textbf{Torsion} & \textbf{Atom Pair} \\ \midrule
\textit{Specialist Models} & & & & & \\
ChemDFM-v2.0-14B & 12.34 & 0.00 & 18.95 & 9.62 & 17.78 \\
InternS1-235B & 18.84 & 0.00 & 26.59 & 18.45 & 28.27 \\ \midrule
\textit{Generalist Models} & & & & & \\
DeepSeek-V3-685B & 15.84 & 0.00 & 20.02 & 14.10 & 22.08 \\
Qwen3-235B & 16.51 & 0.00 & 21.27 & 13.47 & 23.02 \\
KIMI-K2 & 19.72 & 0.50 & 24.73 & 16.94 & 27.75 \\
o3-mini & 18.12 & 1.50 & 22.25 & 14.59 & 24.33 \\
Gemini-2.5-flash & 16.02 & 0.00 & 19.57 & 22.92 & 12.60 \\
Gemini-3-flash & 25.58 & 2.01 & 30.91 & 23.88 & 34.12 \\
GPT-5 & 18.82 & 1.00 & 21.69 & 14.27 & 25.15 \\
Qwen2.5-7B (Base) & 6.56 & 0.00 & 3.93 & 2.63 & 4.16 \\ \midrule
SpecMol-7B w.o. Norm & 50.55 & 25.50 & 53.98 & 46.79 & 55.81 \\
SpecMol-7B (ours) & \textbf{52.44} & \textbf{27.00} & \textbf{55.55} & \textbf{47.62} & \textbf{56.74} \\ \bottomrule
\end{tabular}
\end{table}

%-------------------------------------%

\subsection{Forward Problem: Spectra Generation from SMILES}
\label{sec:smiles2spec}

We next study the forward problem of generating experimental spectra from SMILES across four modalities ($^{13}$C NMR, $^{1}$H NMR, IR, and MS). As summarized in Tab.~\ref{tab:mol2spec} and Tab.~\ref{tab:expert_nmr}, \textit{\mname} consistently delivers the best overall performance among LLM-based baselines, and achieves strong NMR simulation performance under both our primary metrics and NMRSolver-style set scores. Notably, under the reported $^{1}$H NMR set-based metrics, \textit{\mname} attains higher scores than MestReNova (Table~\ref{tab:expert_nmr}), while remaining competitive on $^{13}$C NMR.

\paragraph{Strong and consistent gains across four modalities.}
Tab.~\ref{tab:mol2spec} shows that \textit{\mname} substantially improves both \emph{peak-level} accuracy for NMR and \emph{waveform-level} fidelity for IR/MS. For $^{13}$C NMR, \textit{\mname} raises the F1 score from 0.254 (best LLM baseline) to \textbf{0.479} while reducing the MAE to \textbf{0.149}, indicating improved peak detection and chemical-shift precision. For $^{1}$H NMR, \textit{\mname} achieves the highest Jaccard similarity (\textbf{0.449}) and the lowest MAE (\textbf{0.033}), reflecting more accurate recovery of dense and overlapping proton environments. Beyond discrete peak sets, \textit{\mname} also improves continuous-spectrum similarity, reaching cosine similarities of \textbf{0.554} (IR) and \textbf{0.423} (MS), demonstrating that the model can jointly handle heterogeneous spectral formats within a unified text generation framework.

\paragraph{NMR simulation and comparison to MestReNova.}
To assess chemical realism under stricter NMR metrics, we further evaluate on NMRSolver-style set-based scores (Tab.~\ref{tab:expert_nmr}). \textit{\mname} achieves $S_{set}=\textbf{0.714}$ and $S_{v2v}=\textbf{0.835}$ on $^{13}$C NMR, and $S_{set}=\textbf{0.827}$ and $S_{v2v}=\textbf{0.915}$ on $^{1}$H NMR. Compared with \textbf{MestReNova}~\cite{willcott2009mestre}, \textit{\mname} is competitive on $^{13}$C NMR but \textbf{attains higher scores than MestReNova on $^{1}$H NMR} (0.827 vs. 0.741 in $S_{set}$; 0.915 vs. 0.897 in $S_{v2v}$). This suggests that large-scale multi-source training enables \textit{\mname} to better capture subtle proton chemical environments and coupling-related effects that are challenging for rigid rule-based or single-engine pipelines.

\paragraph{Qualitative case studies.}
Fig.~\ref{fig:performance_overall}F visualizes representative predictions. For a complex aromatic system (left), \textit{\mname} produces $^{13}$C shifts closely aligned with the ground truth (MAE = \textbf{0.1367} ppm), with accurate coverage across both aliphatic and aromatic regions. For IR prediction (right), \textit{\mname} recovers characteristic absorption patterns (MAE = 3.0264), including the prominent carbonyl stretching band near $1700~\mathrm{cm}^{-1}$ and fine-grained structure in the fingerprint region. These examples corroborate the quantitative results and highlight \textit{\mname}'s potential as a practical spectral simulator that bridges molecular topology with experimental observables.

\paragraph{Failure modes and limitations.}
We find that the remaining errors are dominated by experimental and representational uncertainty rather than gross chemical invalidity. In MS/IR generation, the model sometimes produces spectra with correct coarse shape but misses low-intensity diagnostic peaks, which is consistent with the fact that fragmentation pathways and baseline/normalization vary across instruments and protocols. In $^{1}$H NMR, long-range coupling and peak crowding can cause minor deviations even when the predicted set of resonances is largely correct.

\begin{table}[!ht] 
    \centering
    \caption{Performance on \textit{SMILES-to-Spectra} across four modalities. \textbf{Bold} indicates the best among LLM-based models.}
    \label{tab:mol2spec}
    \small
    \setlength{\tabcolsep}{3.5pt}
    \begin{tabular}{l|cc|ccc|c|c}   
    \toprule
    \textbf{Model} & \multicolumn{2}{c|}{\textsuperscript{13}C-NMR} & \multicolumn{3}{c|}{\textsuperscript{1}H-NMR} & \textbf{IR} & \textbf{MS} \\
    \cmidrule(lr){2-3} \cmidrule(lr){4-6} \cmidrule(lr){7-7} \cmidrule(lr){8-8}
     & F1 (↑) & MAE (↓) & Jacc (↑) & F1 (↑) & MAE ↓ & Cos (↑) & Cos (↑) \\
    \midrule
    Deepseek-V3 (685B) & 0.204 & 0.226 & 0.209 & 0.526 & 0.053 & 0.140 & 0.021 \\
    Qwen3-235B & 0.200 & 0.227 & 0.162 & 0.426 & 0.052 & 0.164 & 0.046 \\
    KIMI-K2 & 0.254 & 0.219 & 0.216 & 0.513 & 0.052 & 0.150 & 0.074 \\
    o3-mini & 0.186 & 0.238 & 0.164 & 0.419 & 0.051 & 0.095 & 0.185 \\
    GPT-5-mini & 0.212 & 0.226 & 0.223 & 0.495 & 0.048 & 0.149 & 0.223 \\
    \midrule
    \textbf{SpecMol-7B} & \textbf{0.479} & \textbf{0.149} & \textbf{0.449} & \textbf{0.658} & \textbf{0.033} & \textbf{0.554} & \textbf{0.423} \\
    \bottomrule
    \end{tabular}
\end{table}

\begin{table}[!ht]
    \centering
    \caption{Expert-level comparison on NMR spectral simulation. We report Set Similarity ($S_{set}$) and $Set2Vec$ with Cosine Similarity ($S_{v2v}$). \textbf{Bold} indicates the best among LLM-based models; \underline{underline} indicates where SpecMol outperforms the specialized expert tool.}
    \label{tab:expert_nmr}
    \small
    \setlength{\tabcolsep}{5pt}
    \begin{tabular}{l|cc|cc}
    \toprule
    \multirow{2}{*}{\textbf{Models \textbackslash Metrics}} & \multicolumn{2}{c|}{\textbf{\textsuperscript{13}C-NMR}} & \multicolumn{2}{c}{\textbf{\textsuperscript{1}H-NMR}} \\
    \cmidrule(lr){2-3} \cmidrule(lr){4-5}
     & $S_{set}$ (↑) & $S_{v2v}$ (↑) & $S_{set}$ (↑) & $S_{v2v}$ (↑) \\ \midrule
    \textit{LLM-Based Models} & & & & \\
    DeepSeek-V3 & 0.468 & 0.667 & 0.709 & 0.864 \\
    Qwen3-235B & 0.481 & 0.694 & 0.623 & 0.815 \\
    KIMI-K2 & 0.564 & 0.775 & 0.738 & 0.882 \\
    o3-mini & 0.505 & 0.710 & 0.658 & 0.815 \\
    Gemini-2.5-Flash-lite & 0.402 & 0.576 & 0.629 & 0.787 \\
    GPT-5-mini & 0.526 & 0.727 & 0.728 & 0.875 \\
    \textbf{SpecMol-7B} & \textbf{0.714} & \textbf{0.835} & \textbf{0.827} & \textbf{0.915} \\ \midrule
    \textit{Specialized Tools} & & & & \\
    MestReNova & 0.860 & 0.951 & 0.741 & 0.897 \\
    \textbf{SpecMol-7B} & 0.714 & 0.835 & \underline{\textbf{0.827}} & \underline{\textbf{0.915}} \\ \bottomrule
    \end{tabular}
\end{table}

\begin{table}[!ht]
\centering
\caption{Performance on Molecular Name Conversion. For \textit{SMILES $\to$ IUPAC}, InChIKey is derived via predicted names; for \textit{IUPAC $\to$ SMILES}, RDKit is used for SMILES standardization and InChIKey conversion. \textbf{Bold} indicates the best among LLM-based models.}
\label{tab:name_conversion}
\footnotesize % 字体调小以适应多列
\setlength{\tabcolsep}{2.5pt} % 压缩列间距
\begin{tabular}{l|ccc|ccc}
\toprule
\multirow{2}{*}{\textbf{Model}} & \multicolumn{3}{c|}{\textbf{SMILES $\to$ IUPAC Name}} & \multicolumn{3}{c}{\textbf{IUPAC Name $\to$ SMILES}} \\
\cmidrule(lr){2-4} \cmidrule(lr){5-7}
 & Token (↑) & Seq (↑) & InChIKey (↑) & Token (↑) & Seq (↑) & InChIKey (↑) \\ \midrule
\rowcolor[HTML]{F2F2F2} \multicolumn{7}{l}{\textit{Specialized Tools (Rule-based/Specialist Transformers)}} \\
STOUT v1.0~\cite{rajan2021stout} & 38.38 & 19.00 & 30.00 & 99.59 & 59.00 & 59.00 \\
OPSIN~\cite{lowe2022opsin} & / & / & / & 99.43 & 98.00 & 98.00 \\ \midrule
\rowcolor[HTML]{F2F2F2} \multicolumn{7}{l}{\textit{Task-Specific Specialist Models}} \\
% ChemLLM-7B & 0.10 & 0.00 & 0.00 & 2.26 & 0.00 & 0.00 \\
ChemDFM-v2.0-14B & 41.52 & 20.00 & 38.00 & 78.68 & 78.00 & 78.00 \\
Galactica-6.7B & 1.72 & 0.00 & 0.00 & 5.28 & 1.00 & 1.00 \\
InternS1-235B & 33.73 & 39.00 & 39.00 & 61.68 & 51.00 & 51.00 \\ \midrule
\rowcolor[HTML]{F2F2F2} \multicolumn{7}{l}{\textit{LLM-based Generalist Models}} \\
DeepSeek-V3-685B & 5.81 & 0.00 & 0.00 & 39.07 & 12.00 & 12.00 \\
Qwen3-235B & 4.75 & 0.00 & 0.00 & 15.20 & 0.00 & 0.00 \\
KIMI-K2 & 13.09 & 1.00 & 1.00 & 40.15 & 22.00 & 22.00 \\
o3-mini & 11.94 & 1.00 & 1.00 & 41.19 & 18.00 & 18.00 \\
Gemini-3-Flash & 21.61 & 3.06 & 6.00 & 54.91 & 33.33 & 33.33 \\
GPT-5 & 12.01 & 1.00 & 6.00 & 45.91 & 28.00 & 28.00 \\
Qwen2.5-7B (Base) & 2.94 & 0.00 & 0.00 & 5.95 & 0.00 & 0.00 \\ \midrule
\textbf{\mname-7B} & \textbf{37.64} & \textbf{20.00} & \textbf{29.00} & \textbf{46.07} & \textbf{45.00} & \textbf{45.00} \\ \bottomrule
\end{tabular}
\end{table}

\subsection{Molecular Name Conversion}
\label{sec:results_conversion}

We evaluate \textit{\mname} on bidirectional molecular name conversion between IUPAC names and SMILES, a task that probes whether the model can faithfully translate between natural-language chemical nomenclature and canonicalized molecular graphs (Tab.~\ref{tab:name_conversion}). We report token/sequence accuracy and additionally use \textit{InChIKey accuracy} to measure structure-level correctness. For \textit{IUPAC $\rightarrow$ SMILES}, predicted SMILES are standardized with RDKit before scoring; for \textit{SMILES $\rightarrow$ IUPAC}, we derive InChIKeys by parsing the predicted names and converting them to structures.

\paragraph{SMILES $\rightarrow$ IUPAC: approaching specialized translation tools.}
In the SMILES-to-IUPAC task, \textit{\mname} achieves \textbf{37.64\%} token accuracy and \textbf{20.00\%} sequence accuracy, matching the specialist tool STOUT v1.0 (38.38\% / 19.00\%). More importantly, \textit{\mname} reaches an \textbf{InChIKey accuracy of 29.00\%}, close to STOUT (30.00\%), while generalist LLMs remain far behind (e.g., GPT-5 and Gemini-3-Flash at $\leq$6\% InChIKey accuracy). These results indicate that \textit{\mname} learns not only the surface form of IUPAC grammar but also preserves structure-level semantics during generation.

\paragraph{IUPAC $\rightarrow$ SMILES: narrowing the gap to rule-based parsers.}
For the reverse direction, deterministic parsers such as OPSIN remain strong due to handcrafted rules (98.00\% InChIKey). Nevertheless, \textit{\mname} achieves \textbf{45.00\%} sequence accuracy and \textbf{45.00\%} InChIKey accuracy, outperforming all generalist LLM baselines and bridging a substantial portion of the gap toward specialized molecular models. Compared to its base model (Qwen2.5-7B), which fails on nearly all exact-matching cases, this improvement highlights the benefit of our structure-grounded pretraining and multi-task SFT in learning the near-bijective mapping between systematic names and 2D topology.

\begin{figure}[!htb]
    \centering
    \includegraphics[width=\linewidth]{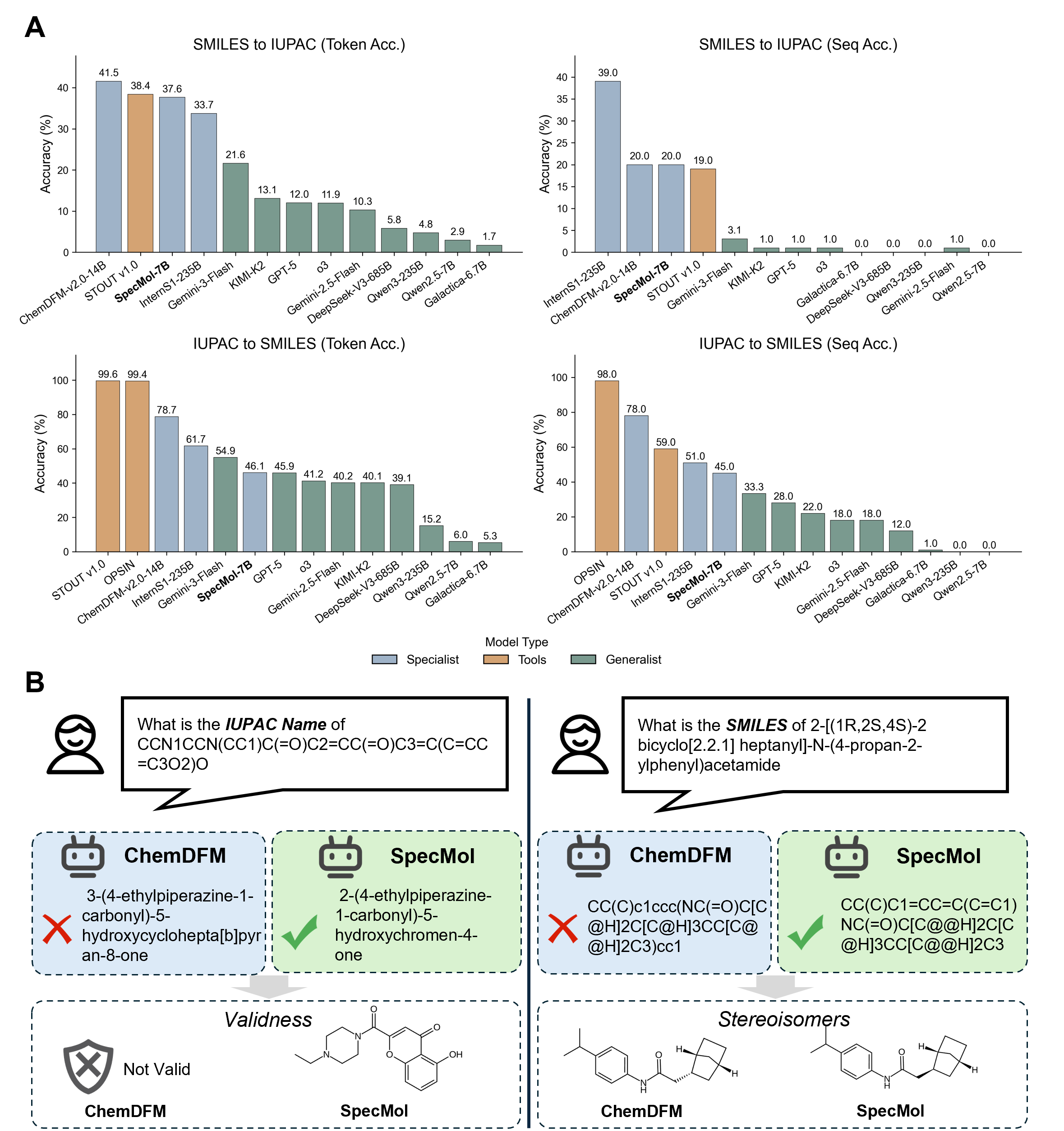}
    \caption{Quantitative and qualitative results of SMILES-IUPAC bidirectional translation. (A) Accuracy metrics across various models, categorized by model type (Specialist, Tools, Generalist). SpecMol consistently outperforms generalist LLMs and matches specialized chemical tools. (B) Comparative case studies on structural validity and stereoisomer identification. SpecMol correctly decodes complex nomenclature and stereochemical configurations where previous specialist models exhibit structural invalidity or stereochemical errors.}
    \label{fig:smiles_iupac_cases}
\end{figure}

\paragraph{Qualitative analysis: reducing hallucinations and improving stereochemical fidelity.}
Fig.~\ref{fig:smiles_iupac_cases}B highlights two representative examples comparing \textit{\mname} against ChemDFM. In the first case, ChemDFM hallucinates an incorrect core scaffold, whereas \textit{\mname} recovers the correct chromen-4-one nucleus and substituents. In the second case featuring multiple chiral centers, \textit{\mname} more reliably preserves stereochemical assignments when mapping $R/S$ descriptors into SMILES chirality markers (\texttt{@}/\texttt{@@}), suggesting stronger alignment between textual nomenclature and graph-level stereochemistry.

\subsection{Molecule QA}
\label{sec:results_qa}

We evaluate \textit{\mname} on Molecule QA (MolQA) to assess general chemical reasoning beyond direct translation, covering questions that require connecting molecular structure to properties and functional relationships (Fig.~\ref{fig:performance_overall}D). \textit{\mname} achieves an accuracy of \textbf{68.6\%}, substantially improving over its base model Qwen2.5-7B (32.0\%) and outperforming much larger open generalist models such as DeepSeek-V3-685B (54.6\%) and Qwen3-235B (52.0\%), as well as specialist ChemDFM-v2.0-14B (59.6\%).

We attribute these gains to spectroscopy-grounded pretraining and multi-task alignment, which strengthen the model’s internal representation of functional groups and local chemical environments---signals that are directly relevant to both spectral interpretation and property-centric QA. Nevertheless, \textit{\mname} still trails proprietary frontier models (e.g., GPT-5 and o3), suggesting that further improvements may require larger-scale instruction data and broader chemical knowledge integration.

\subsection{3D Structure Generation}
\label{sec:results_3d}

We finally evaluate \textit{\mname} on direct 3D coordinate generation from symbolic inputs (SMILES or IUPAC), which requires the model to jointly preserve chemical connectivity and produce physically plausible geometries. As shown in Tab.~\ref{tab:coord-bench}, \textit{\mname} achieves the highest validity on both SMILES-to-3D and IUPAC-to-3D, demonstrating strong robustness compared to substantially larger generalist LLMs.

\begin{table}[!htb]
    \centering
    \caption{Performance on 3D coordinate generation tasks. \textbf{Bold} indicates the best performance; \underline{Underline} highlights SpecMol's leading validity.}
    \label{tab:coord-bench}
    \footnotesize % 缩小字号以适应 9 列数据
    \setlength{\tabcolsep}{2.5pt} % 压缩列间距
    \begin{tabular}{l|cccc|cccc}
    \toprule
    \multirow{2}{*}{\textbf{Model}} & \multicolumn{4}{c|}{\textbf{SMILES-to-3D}} & \multicolumn{4}{c}{\textbf{IUPAC-to-3D}} \\
    \cmidrule(lr){2-5} \cmidrule(lr){6-9}
      & Valid (↑) & Clash (↓) & Viol (↓) & FP (↑) & Valid (↑) & Clash (↓) & Viol (↓) & FP (↑) \\
    \midrule
    Deepseek-V3 (685B) & 16.50 & 8.941 & 0.151 & 0.152 & 42.50 & \textbf{1.138} & 3.543 & \textbf{0.721} \\
    KIMI-K2 & 22.00 & \textbf{0.000} & \textbf{0.000} & 0.315 & 11.50 & 1.375 & 4.750 & 0.574 \\
    o3-mini & 45.50 & 1.825 & 2.175 & 0.356 & 54.50 & 5.027 & 4.186 & 0.642 \\
    Gemini-2.5-Flash & 64.00 & 5.672 & 2.270 & 0.304 & 63.50 & 52.19 & 3.810 & 0.693 \\
    GPT-5 & 69.50 & 0.224 & 0.217 & 0.314 & 53.00 & 4.686 & 2.059 & 0.813 \\
    \midrule
    \textbf{\mname (7B)} & \underline{\textbf{89.68}} & 2.880 & 0.994 & \textbf{0.582} & \underline{\textbf{82.78}} & 3.012 & \textbf{1.357} & 0.705 \\
    \bottomrule
    \end{tabular}
\end{table}

\begin{figure}[!ht]
    \centering
    \includegraphics[width=0.8\linewidth]{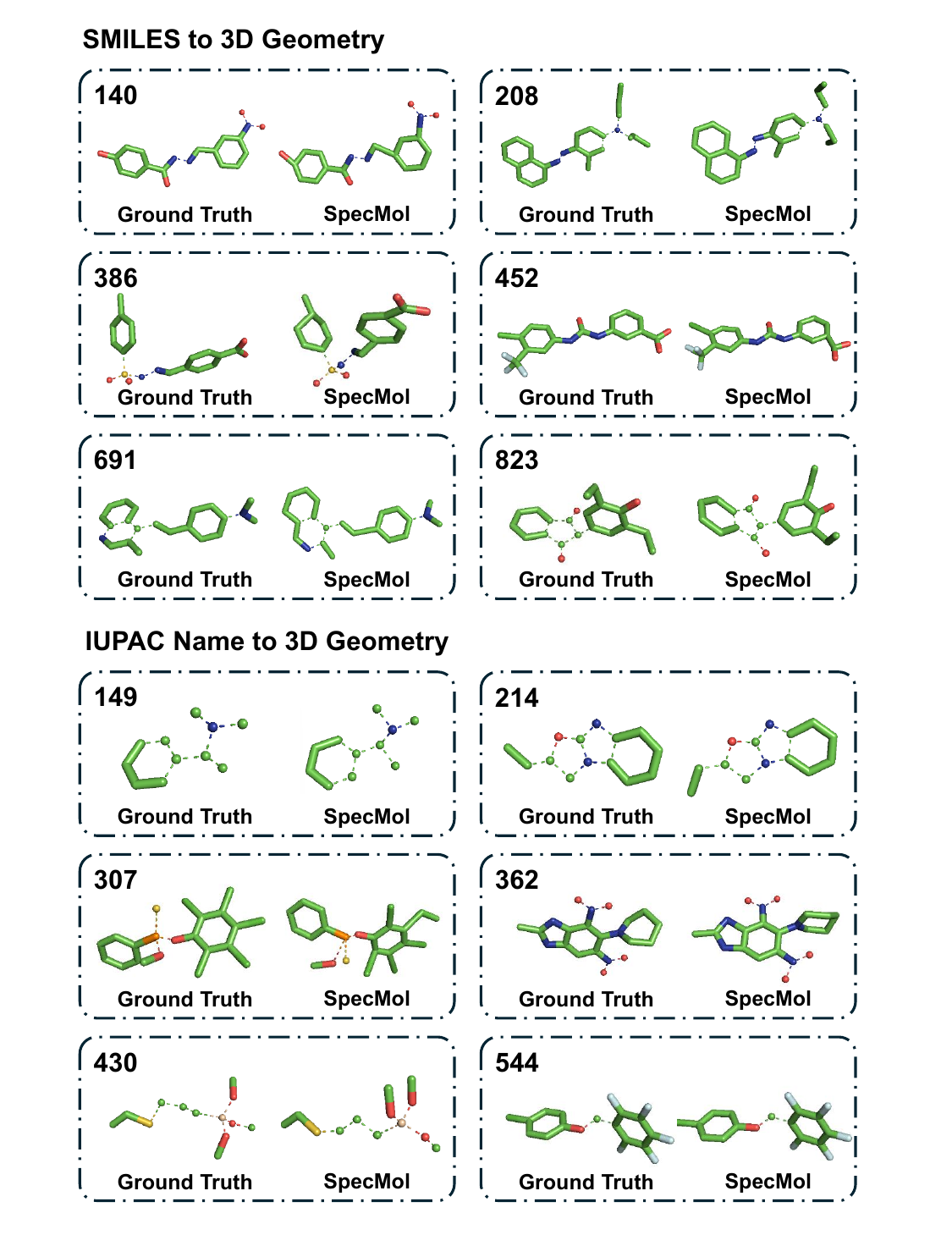}
    \caption{Visual comparison of 3D molecular structures generated by \mname~against the Ground Truth. The upper panel shows structures generated from \textbf{SMILES}, and the lower panel shows structures generated directly from \textbf{IUPAC names}. \mname~accurately recovers complex geometries, including ring systems, stereocenters, and spatial orientations of functional groups, demonstrating high geometric fidelity across diverse molecular scaffolds.}
    \label{fig:3d_visual_comparison}
\end{figure}

\paragraph{Quantitative geometric accuracy and robustness.}
On SMILES-to-3D, \textit{\mname} attains the best \textit{SDF validity} of \textbf{89.68\%} and the highest fingerprint similarity (\textbf{0.582}), indicating that generated conformations are not only parsable but also strongly consistent with the intended 2D connectivity. While some baselines exhibit lower clash scores, they typically suffer from much lower validity (e.g., KIMI-K2: 22.00\% validity), suggesting a robustness--geometry trade-off where producing valid structures at scale remains the primary bottleneck. On the more challenging IUPAC-to-3D setting, \textit{\mname} preserves high validity (\textbf{82.78\%}) and achieves the lowest bond violation (\textbf{1.357}), demonstrating improved physical plausibility when directly grounding systematic nomenclature into 3D coordinates. Although GPT-5 reports a higher FP similarity (0.813) on IUPAC-to-3D, its substantially lower validity (53.00\%) indicates weaker end-to-end robustness, whereas \textit{\mname} provides consistently parsable and physically reasonable outputs across a wider portion of the test set.

\paragraph{Qualitative geometric fidelity.}
Fig.~\ref{fig:3d_visual_comparison} further confirms that \textit{\mname} captures complex spatial arrangements. For both SMILES (Rows 1--3) and IUPAC (Rows 4--6) inputs, predicted conformations (green) show strong overlap with ground-truth experimental structures across diverse scaffolds. Specifically, \textit{\mname} recovers (i) ring planarity and relative orientations in rigid aromatic and amide-containing systems (e.g., cases 140 and 452), (ii) correct stereochemical configurations around chiral centers (e.g., 362 and 430), and (iii) plausible geometries for challenging bicyclic and branched molecules (e.g., 214 and 823). These results suggest that treating 3D coordinates as a first-class supervision target during SFT enables \textit{\mname} to move beyond text-only generation toward geometry-aware molecular modeling.

%% file: files/conclusion.tex
\section{Conclusion}
% We introduce \textsc{\mnametwo}, a unified benchmark and chemistry-aware evaluation protocol for assessing LLMs across spectra $\leftrightarrow$ structure elucidation, structure $\rightarrow$ spectra simulation, SMILES/IUPAC translation, molecular QA, and SMILES  $\rightarrow$ 3D conformation generation under consistent splits, standardized serialization, and reproducible decoding/post-processing. \textsc{\mnametwo} addresses the fragmentation of prior evaluation settings by providing modality-native textual representations (tagged, fixed-point serialization) and a unified metric suite spanning symbolic correctness, spectral fidelity, and physical plausibility. Using \textit{\mname} as a controlled reference baseline, we show that model rankings and reported performance can vary substantially under inconsistent protocols, and that several strong LLMs underperform when evaluated with laboratory-grounded, standardized settings. Overall, \textsc{\mnametwo} establishes reproducible standards to guide the development of spectroscopy-grounded molecular AI that connects experimental observables with structural and geometric reasoning. Future work will focus on expanding benchmark coverage across instruments and conditions, strengthening instruction adherence and structured output reliability, and improving geometry-/physics-aware generation (e.g., reducing clashes and bond violations) while maintaining robustness and validity.

We introduce \textsc{\mnametwo}, a foundation model capable of precise cross-modal reasoning between molecular structures and spectroscopic data, together with SpecMol-Bench, a unified evaluation framework designed to standardize assessment across spectroscopy-grounded tasks—from spectra-to-structure elucidation to 3D conformation generation.
SpecMol-Bench addresses the fragmentation of prior evaluation settings by providing modality-native textual representations and a cohesive metric suite that jointly assesses symbolic correctness, spectral fidelity, and physical plausibility. Using SpecMol as a controlled reference, our extensive experiments reveal two critical findings: (1) model rankings can vary substantially under inconsistent protocols, and (2) several strong generalist LLMs underperform when evaluated in laboratory-grounded settings.
Together, this work establishes reproducible standards for developing molecular AI that bridges experimental observables with structural and geometric reasoning. Future directions include expanding benchmark coverage across instruments and experimental conditions, improving instruction adherence and structured output reliability, and advancing physically grounded 3D generation—specifically by reducing steric clashes and bond violations while preserving robustness and chemical validity.

%% file: files/contributions.tex
\section{Contributions}
Shuaike Shen is responsible for data curation, \textit{\mname}~model training, and drafting the manuscript. 
Jiaqing Xie is responsible for drafting the manuscript, visualization, and computational experiments. 
Zhuo Yang is responsible for drafting the manuscript,  and computational experiments. 
Shuzhou Sun is responsible for visualization. 
Ben Gao is responsible for drafting the
manuscript, and data curation. 
Tianfan Fu is responsible for drafting the manuscript. 
Biqing Qi is responsible for drafting the manuscript.
Yuqiang Li is responsible for supporting the whole project.

%% file: files/Conflict_Statement.tex
\section{Conflict of Interest Statement}
The authors declare no competing financial interests.

%% file: files/acknowledgement.tex
\section{Acknowledgments}
Tianfan Fu is supported by Nanjing University International Collaboration Initiative and Distinguished Overseas Young Talents.

%% file: files/software.tex
\section{Data and Software Availability Statement
}
Code and partial data is available at \href{https://github.com/Eurekashen/SpecMol}{https://github.com/Eurekashen/SpecMol}.

%% file: files/toc.tex
\section{Table of Contents (TOC) Graphic}
\includegraphics[width=\linewidth]{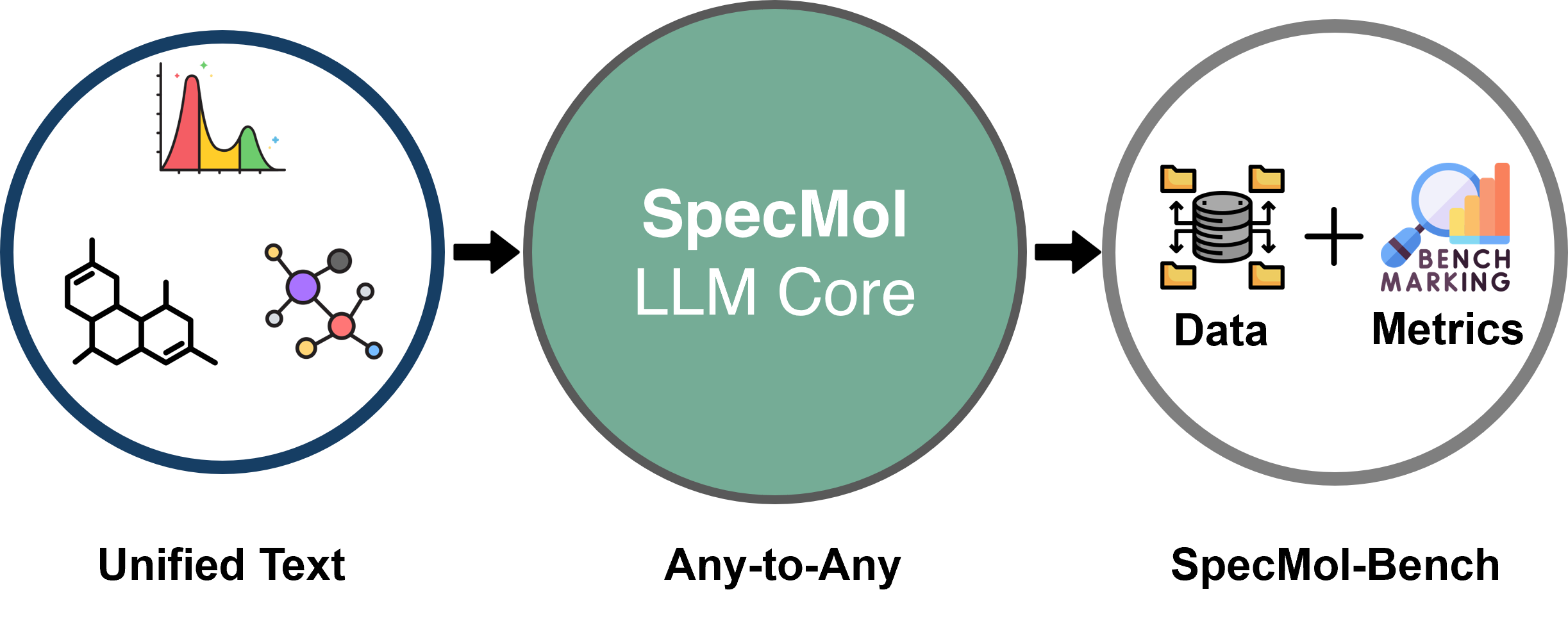}